\documentclass[runningheads]{llncs}
\usepackage[T1]{fontenc}
\usepackage{graphicx}
\usepackage{booktabs}  
\usepackage{multirow}  
\usepackage{colortbl}  
\usepackage{arydshln}  
\usepackage{verbatim}
\usepackage{amsmath,amssymb}
\usepackage{amsfonts}
\usepackage{color}
\usepackage[colorlinks=true,
            linkcolor=blue,
            citecolor=blue,
            urlcolor=blue]{hyperref}
\usepackage{setspace}
\begin{document}
\title{\textsf{DepthPolyp}: Pseudo-Depth Guided Lightweight Segmentation for Real-Time Colonoscopy}
%
\titlerunning{\textsf{DepthPolyp}}
%

\author{Zhuoyu Wu\inst{1}\and
Wenhui Ou\inst{2}\and
Lexi Zhang\inst{4}\and
Pei-Sze Tan\inst{1}\and
Dongjun Wu\inst{2}\and
\newline Junhe Zhao\inst{3}\and
Wenqi Fang\inst{3}\and
Rapha{\"e}l C.-W. Phan\inst{1}}
\authorrunning{Z. Wu et al.}
%
\institute{CyPhi AI Lab, Monash University, Malaysia Campus, Malaysia \and
Department of Electronic \& Computer Engineering, Hong Kong University of Science \& Technology, Hong Kong, P.R. China \and
Shenzhen Institutes of Advanced Technology, Chinese Academy of Sciences, Shenzhen, P.R. China \and
Harbin Institute of Technology, Harbin, P.R. China \\}


%
\maketitle              
\begin{abstract}
Accurate polyp segmentation in colonoscopy is essential for early colorectal cancer detection, yet real-world clinical environments pose persistent challenges such as motion blur, specular reflections, and illumination instability. Most existing methods are optimized on clean benchmark images and suffer noticeable performance degradation when deployed in authentic surgical scenarios.
We propose \textsf{DepthPolyp}, a segmentation framework designed for robustness and lightweight under diverse degradations through pseudo-depth-guided multi-task learning and efficient feature modulation. The architecture combines hierarchical Ghost factorization for compact feature generation, Interleaved Shuffle Fusion for low-cost cross-scale interaction, and Dynamic Group Gating for group-wise adaptive feature weighting.
Extensive experiments demonstrate that \textsf{DepthPolyp} achieves strong cross-dataset generalization when trained on degraded data and evaluated on both clean and noisy target domains, consistently outperforming lightweight baselines and remaining competitive with substantially larger models. In real surgical video evaluation on PolypGen, \textsf{DepthPolyp} attains better segmentation performance than models by up to 20× larger while preserving real-time performance. With only 3.57M parameters and 0.86 GMACs, the proposed method runs at over 180 FPS on mobile devices, making it well suited for real-time deployment in resource-constrained clinical environments. The code and weight can be found at \url{https://github.com/ReaganWu/DepthPolyp/}.

\keywords{Lightweight endoscopy segmentation\and Pseudo depth guidance\and Multi-task learning \and Robustness.}

\end{abstract}
\section{Introduction}
Real-time semantic understanding in medical imaging remains a key challenge for deploying computer-aided diagnosis systems in clinical practice~\cite{mei2025survey}. 
Although recent convolutional and transformer-based architectures achieve high accuracy on curated benchmarks~\cite{hassan2024computer}, their performance degrades sharply under motion blur, specular reflections, and illumination variations commonly observed in endoscopic video streams~\cite{wang2025agentpolyp,jain2024comparative}. 
This discrepancy between controlled evaluation and surgical reality severely limits the practical deployment of polyp segmentation systems, where unstable predictions may directly affect clinical outcomes~\cite{11461918}.

Existing polyp segmentation methods can be broadly grouped into several categories, each with inherent limitations. 
Transformer-based models~\cite{fan2020pranet,li2026cfformer,chen2021transunet} deliver strong performance on clean images but typically require over 30M parameters and show pronounced robustness degradation under blur, exceeding a 20\% Dice drop in our experiments. 
Lightweight models~\cite{karmakar2022mobilepolyp,dinh2023ulite,tang2024cmunext} emphasize efficiency, yet often suffer from limited representation capacity, leading to unstable predictions on degraded inputs. 
Multi-task approaches~\cite{sun2020saunet} introduce auxiliary objectives such as edge or saliency supervision. However, these cues are themselves sensitive to appearance corruption, providing limited robustness improvement in practice. 
More critically, \textbf{most prior works evaluate colonoscopy segmentation models only on high-quality test sets, overlooking degradations such as motion blur and reflection that are prevalent in real procedures}, resulting in an overestimation of real-world reliability.

To address this gap, we propose \textsf{DepthPolyp}, a pseudo-depth-guided framework explicitly designed for robustness under surgical degradations. 
Recent studies have shown that monocular depth estimation encodes structural cues that are less sensitive to appearance corruption~\cite{depthanything3,depth_anything_v2}. 
Motivated by this observation, we leverage pseudo-depth as auxiliary supervision to regularize feature learning during training, rather than enforcing explicit geometric reasoning at inference time. 
Depth-Anything v2~\cite{depth_anything_v2} is used to generate pseudo-depth targets, which are incorporated into a lightweight hierarchical decoder with efficient multi-scale feature fusion and dynamic group-wise modulation. 
An uncertainty-aware multi-task loss~\cite{kendall2018multi} is adopted to automatically balance segmentation and depth supervision, improving robustness without introducing inference-time overhead.

Our main contributions are summarized as follows:
\textbf{(1)} A robustness-oriented evaluation protocol with four configurations (Clean$\rightarrow$Clean, Clean$\rightarrow$Noisy, and Noisy$\rightarrow$Clean, Noisy$\rightarrow$Noisy) across Kvasir~\cite{jha2019kvasir}, CVC-ClinicDB~\cite{bernal2015clinicdb}, and CVC-ColonDB~\cite{bernal2012colondb}, together with authentic surgical degradation sequences from PolypGen~\cite{ali2023polypgen}, exposing performance gaps overlooked by standard benchmarks.
\textbf{(2)} A lightweight segmentation architecture (3.57M parameters, 0.86 GMACs) achieving real-time performance (181 FPS on iPhone 15) while consistently outperforming larger baselines under realistic degradations, including a 9.1\% average Dice improvement over SegFormer-B0~\cite{xie2021segformer}.
\textbf{(3)} A pseudo-depth-guided and uncertainty-aware training strategy that improves robustness without increasing inference-time complexity.
\textbf{(4)} Comprehensive ablation studies validating the contribution of each component, with particular emphasis on uncertainty-aware optimization and dynamic gating.

\section{Related Work}

\noindent
\textbf{Polyp Segmentation Architectures.}
Since the introduction of U-Net~\cite{ronneberger2015u}, encoder--decoder architectures with skip connections have become the dominant paradigm for polyp segmentation. Subsequent works extend this design by incorporating attention mechanisms or transformer components, such as PraNet~\cite{fan2020pranet}, TransUNet~\cite{chen2021transunet}, and SegFormer~\cite{xie2021segformer}, achieving strong performance on standard benchmarks. Several recent models further explore boundary enhancement~\cite{sun2020saunet}, structured embeddings~\cite{xiao2024ctnet}, or hybrid CNN--Transformer designs to improve lesion awareness~\cite{li2026cfformer}. Despite these advances, most methods are developed and evaluated on clean datasets, while the robustness of polyp segmentation models under realistic colonoscopy degradations, including motion blur, illumination variation, and specular artifacts, remains insufficiently studied.

\noindent
\textbf{Lightweight Segmentation Models.}
To enable deployment on resource constrained devices, lightweight architectures for segmentation have been actively explored~\cite{wu2024harmonizing,tang2024cmunext}. Mobile-PolypNet adopts MobileNet-style bottlenecks for efficient feature extraction~\cite{karmakar2022mobilepolyp}, ULite employs axial convolutions to reduce complexity, while MedT~\cite{valanarasu2021medical}, UNeXt~\cite{valanarasu2022unext}, and CMUNeXt~\cite{tang2024cmunext} design compact attention or convolutional modules to balance accuracy and efficiency. Although these methods successfully reduce parameter count and FLOPs, their evaluation is largely limited to high-quality images. The impact of realistic degradations and real-time constraints in surgical scenarios is rarely considered, leaving a gap between lightweight design and practical robustness.

\noindent
\textbf{Depth-Guided Segmentation.}
Monocular depth estimation provides complementary geometric cues that can benefit semantic segmentation when used as auxiliary supervision~\cite{zhu2020edgedepth}. Multi-task frameworks such as SwinMTL~\cite{taghavi2024swinmtl} and ADRNet-S~\cite{zhou2024adrnet} exploit shared representations for depth and segmentation, while EdgeDepth introduces explicit depth-based constraints~\cite{zhu2020edgedepth}. In polyp segmentation, recent works leverage pseudo-depth from pretrained depth estimators to guide compact models~\cite{phuong2024bbd,zheng2024polyp}. However, these approaches primarily focus on accuracy gains under clean conditions and mainly for semi-supervised learning. Which provides a limited analysis of robustness, inference efficiency, and out-of-distribution validation, which are critical for real-world colonoscopy deployment.

In contrast to existing studies, our work investigates depth-guided learning from a robustness-first perspective. We focus on lightweight architecture design, degradation-aware evaluation, and real-time inference, aiming to bridge the gap between benchmark performance and practical clinical applicability.

\begin{figure}[t]
\centering
\includegraphics[width=1.0 \textwidth]{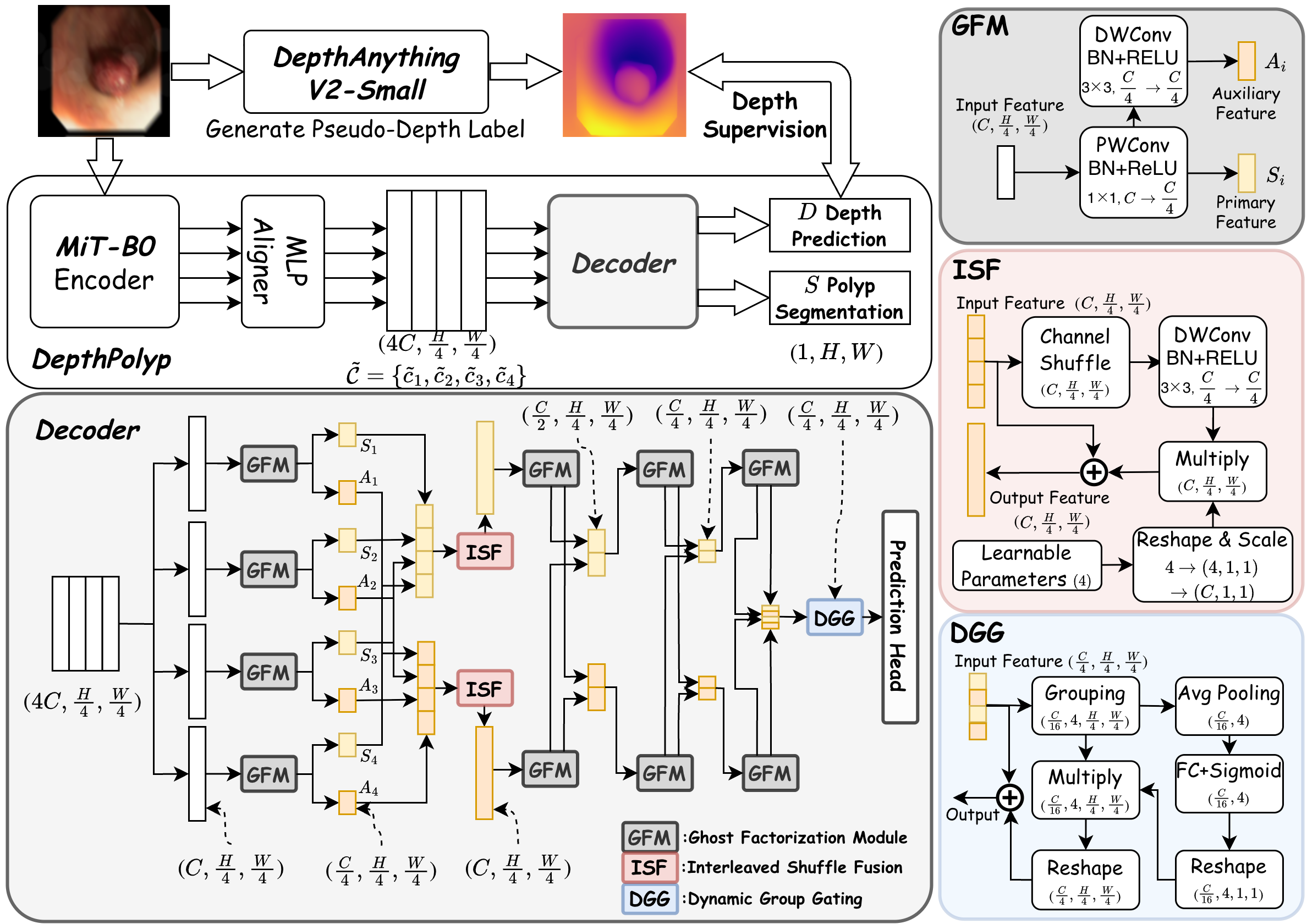}
\caption{
Overview of the proposed \textsf{DepthPolyp} framework.
During training (upper-left), the input image is processed by \textsf{DepthPolyp}
together with a frozen Depth-Anything v2 (Small) model to provide pseudo-depth
supervision.
\textsf{DepthPolyp} jointly predicts segmentation and auxiliary depth, while
pseudo-depth is used only during training to encourage geometry-aware learning.
The lightweight decoder (lower-left) integrates Ghost Factorization (GFM),
Interleaved Shuffle Fusion (ISF), and Dynamic Group Gating (DGG) for efficient
multi-scale feature aggregation.
The right panels depict the structures of GFM, ISF, and DGG
(details in Sec.~\ref{sec:method}).
}
\label{fig:overview}
\end{figure}

\section{Method}
\label{sec:method}

We present \textsf{DepthPolyp}, a lightweight polyp segmentation framework designed for robustness under surgical degradations through (1) computationally factorized feature representations, (2) efficient high-resolution interleaved fusion, and (3) depth-guided uncertainty-aware multi-task learning. The overall pipeline is shown in Fig.~\ref{fig:overview}.

\subsection{Notation and overview}
\label{sec:notation}
Given an input image batch \(I\in\mathbb{R}^{B\times 3\times H\times W}\),
an MiT-B0 encoder produces four multi-scale feature maps
\(\{c_1, c_2, c_3, c_4\}\) with progressively reduced spatial resolutions.
Each feature is projected to a unified channel dimension via a token-wise
linear layer, reshaped to the spatial domain, and upsampled to a common
resolution of \(H/4\times W/4\):
\begin{equation}
\tilde{c}_i = \mathrm{Upsample}\big(\mathrm{reshape}(\mathrm{MLP}_i(c_i)),\;
\text{size}=(H/4,W/4)\big).
\end{equation}
The decoder aggregates \(\tilde{\mathcal C}=\{\tilde c_1,\tilde c_2,\tilde c_3,\tilde c_4\}\)
into a fused representation
\(F_{\mathrm{out}}\in\mathbb{R}^{B\times C_{\mathrm{out}}\times H/4\times W/4}\),
from which segmentation and depth outputs are generated.

\subsection{Ghost Factorization Module (GFM)}
\label{sec:method-gfm}
GFM is inspired by the GhostNet principle of generating redundant feature maps through cheap operations~\cite{han2020ghostnet}, but is adapted here as a hierarchical decoder factorization for dense segmentation. Specifically, we decompose features into a primary component implemented by pointwise transformation and an auxiliary component generated by a cheaper depthwise operation. This design enables computational decomposition rather than explicit semantic disentanglement.

Given input \(X\in\mathbb{R}^{B\times C_{in}\times H\times W}\), the GFM computed as:
\begin{align}
X_p &= \mathrm{PWConv}(X)  \in\mathbb{R}^{B\times C_p\times H\times W},\label{eq:gfm_pw}\\
X_a &= \mathrm{DWConv}(X_p) \in\mathbb{R}^{B\times C_a\times H\times W},\label{eq:gfm_dw}
\end{align}
where \(\mathrm{PWConv}\) is a pointwise \(1\times1\) convolution (with BN and ReLU) and \(\mathrm{DWConv}\) is a depthwise spatial convolution (with BN and ReLU). The GFM returns the two outputs:
\[
\text{GFM}(X) = \big(X_p,\; X_a\big),
\]
with \(C_p + C_a = C_{\text{out\_GFM}}\). In practice we set a split ratio \(r\) (e.g. \(r=2\)) so that \(C_p \approx C_{\text{out\_GFM}}/r\). The composition approximates a full dense convolution with significantly fewer parameters.

\subsection{Hierarchical Factorized Decoder}
\label{sec:method-decoder}
The decoder applies GFM hierarchically in three stages to enable efficient
multi-scale aggregation.

\paragraph{Stage I: per-scale factorization.}
Each unified feature \(\tilde c_i\) is factorized independently:
\begin{equation}
(S_i, A_i) = \text{GFM}(\tilde c_i),\qquad i=1,\dots,4,
\end{equation}
where \(S_i\) and \(A_i\) denote the primary and auxiliary components,
respectively. We form concatenated primary and auxiliary streams across scales:
\begin{equation}
\mathcal S_1 = [S_4, S_3, S_2, S_1],\qquad \mathcal A_1 = [A_4, A_3, A_2, A_1],
\end{equation}
where \([\cdot]\) denotes channel-wise concatenation.

\paragraph{Stage II: cross-stream refinement.}
Both streams are fused using Interleaved Shuffle Fusion (ISF,
Sec.~\ref{sec:method-isf}) for low-cost cross-scale interaction, followed
by another GFM to further compress and refine the representations:
\[
\big(S_{\mathcal S}, A_{\mathcal S}\big)=\mathrm{GFM}\big(\mathrm{ISF}(\mathcal S_1)\big),\quad
\big(S_{\mathcal A}, A_{\mathcal A}\big)=\mathrm{GFM}\big(\mathrm{ISF}(\mathcal A_1)\big).
\]

\paragraph{Stage III: adaptive aggregation.}
All refined components are concatenated and adaptively modulated using
Dynamic Group Gating (DGG, Sec.~\ref{sec:method-dgg}):

\begin{equation}
F_{\mathrm{out}} = \mathrm{DGG}\big([S_{\mathcal S}, S_{\mathcal A}, A_{\mathcal S}, A_{\mathcal A}]\big).
\end{equation}
Empirically we set \(C_{\mathrm{out}}=64\) for the final fused feature.

\subsection{Interleaved Shuffle Fusion (ISF)}
\label{sec:method-isf}
Interleaved Shuffle Fusion (ISF) enables lightweight cross-group interaction
through deterministic channel shuffling followed by spatial refinement.
Given an input feature \(F \in \mathbb{R}^{B \times C \times H \times W}\),
channels are evenly divided into \(G\) groups (\(G=4\) in this work).

First, a fixed channel shuffle operator $\mathrm{Shuffle}_G(\cdot)$ interleaves channels across groups, producing:
\begin{equation}
\hat{F} = \mathrm{Shuffle}_G(F),
\end{equation}
which promotes information exchange without introducing parameters.

Second, spatial refinement is applied using a depthwise convolution:
\begin{equation}
U = \mathrm{DWConv}(\hat{F}).
\end{equation}

Finally, group-wise learnable scales $\gamma \in \mathbb{R}^{G}$ are broadcast to the channel dimension and applied in a residual manner:
\begin{equation}
F' = F + \mathrm{expand}(\gamma) \odot U.
\end{equation}
ISF introduces minimal overhead while facilitating efficient multi-scale
feature interaction.

\subsection{Dynamic Group Gating (DGG)}
\label{sec:method-dgg}
Dynamic Group Gating (DGG) performs group-wise adaptive feature modulation.
The input feature is reshaped along the channel dimension into explicit
groups:
\begin{equation}
\tilde{X} \in \mathbb{R}^{B \times G \times C_g \times H \times W}, 
\quad C_g = C / G.
\end{equation}

A group descriptor is obtained by average pooling over the channel and spatial dimensions:
\begin{equation}
z = \mathrm{AvgPool}(\tilde{X}) \in \mathbb{R}^{B \times G}.
\end{equation}

Group-wise gates are predicted via a lightweight linear projection:
\begin{equation}
w = \sigma(\phi(z)) \in (0,1)^{B \times G}.
\end{equation}

The gated feature is computed by broadcasting $w$ along $(C_g, H, W)$:
\begin{equation}
\tilde{X}' = \tilde{X} \odot w^{\uparrow},
\end{equation}
which is then reshaped back to the original layout and added residually:
\begin{equation}
X_{\text{out}} = X + \mathrm{Reshape}(\tilde{X}').
\end{equation}
The gated features are reshaped back to the original layout and added
residually to preserve the original representation.

\subsection{Depth-guided multi-task learning}
\label{sec:method-loss}

The network is trained in a multi-task manner with two prediction heads: a segmentation head producing logits
$S_{logit}\in\mathbb{R}^{B\times 1\times H\times W}$, and a depth head predicting a normalized depth map
$D\in\mathbb{R}^{B\times 1\times H\times W}$.

\paragraph{Segmentation loss.}
The segmentation output is activated by a sigmoid function to obtain the probability map
$p=\sigma(S_{logit})$.
We adopt the Dice loss as the sole segmentation objective due to its robustness to foreground–background imbalance:
\begin{equation}
\mathcal{L}_{\mathrm{seg}}
=
\mathcal{L}_{\mathrm{Dice}}
=
1 - \frac{2\sum p\, y + \epsilon}{\sum p + \sum y + \epsilon},
\end{equation}
where \(y\) denotes the ground-truth mask and \(\epsilon\) is a small constant for numerical stability.

\paragraph{Depth loss.}
Since the pseudo-depth from Depth Anything v2 represents relative depth
rather than metric scale, predictions are constrained to the range \([0,1]\).
Accordingly, depth labels are normalized to the same range during training.
We apply the Smooth-\(L_1\) loss for depth regression:
\begin{equation}
\mathcal{L}_{\mathrm{depth}} = \mathrm{SmoothL1}(D, D^{*}),
\end{equation}
where \(D^{*}\) denotes the normalized pseudo-depth supervision.

\paragraph{Uncertainty-weighted joint optimization.}
To balance the segmentation and depth objectives without manual tuning, we adopt uncertainty-based weighting:
\begin{equation}
\mathcal{L}
=
\frac{1}{2\sigma_s^2}\mathcal{L}_{\mathrm{seg}}
+
\frac{1}{2\sigma_d^2}\mathcal{L}_{\mathrm{depth}}
+
\log\sigma_s + \log\sigma_d,
\end{equation}
where \(\sigma_s\) and \(\sigma_d\) are learnable task uncertainty parameters.
This formulation enables automatic balancing between segmentation accuracy
and depth consistency during training.


\section{Experiments}
We conduct extensive experiments to evaluate the robustness, accuracy, and efficiency of \textsf{DepthPolyp} across multiple endoscopic datasets and challenging blur/noise conditions. 
\subsection{Datasets}

We conduct experiments on four widely-used polyp segmentation datasets to evaluate our method's performance and generalization capability. \textbf{Kvasir-SEG}~\cite{jha2019kvasir} provides 1,000 high-quality polyp images with pixel-level annotations and serves as our primary training set. For cross-domain evaluation, we employ \textbf{CVC-ClinicDB}~\cite{bernal2015clinicdb} (612 images) and \textbf{CVC-ColonDB}~\cite{bernal2012colondb} (380 images), which contain diverse polyp appearances and imaging conditions. To assess real-world domain generalization under challenging surgical scenarios, we utilize sequences 18, 19, 20, 21, and 22 from \textbf{PolypGen}~\cite{ali2023polypgen} (273 images total), which specifically capture adverse conditions including motion blur and severe reflection artifacts commonly encountered during clinical procedures. Table~\ref{tab:datasets} summarizes the datasets used in this study.

\begin{table}[h]
\scriptsize
\centering
\caption{Summary of datasets used in this study. Train \& Val: The Kvasir-SEG dataset is split into 80\% for training and 20\% for validation. OOD Val: Out-of-distribution (OOD) validation is performed using the weights trained on Kvasir-SEG and evaluated on datasets that are unseen by the model.}
\label{tab:datasets}
\begin{tabular}{lccc}
\toprule
\textbf{Dataset} & \textbf{\# Images} & \textbf{Usage} & \textbf{Characteristics} \\
\midrule
Kvasir-SEG~\cite{jha2019kvasir} & 1,000 & Train\&Val & High-quality polyp images \\
CVC-ClinicDB~\cite{bernal2015clinicdb} & 612 & OOD Val & Cross-domain evaluation \\
CVC-ColonDB~\cite{bernal2012colondb} & 380 & OOD Val & Diverse polyp appearances \\
PolypGen(Seq. 18-22)~\cite{ali2023polypgen} & 273 & OOD Val & Surgical scenario with blur \\
& & & and reflection artifacts \\
\bottomrule
\end{tabular}
\end{table}

\subsection{Robustness-Oriented Degradation Synthesis}
\label{sec:degradation}

Training on clean Kvasir-SEG images alone causes severe performance collapse under real surgical conditions (Table~\ref{tab:robustness}). To bridge this gap, we apply synthetic degradations to both training data and OOD test sets (CVC-ClinicDB, CVC-ColonDB), creating clean-noisy evaluation pairs that enable our four-quadrant robustness protocol. PolypGen sequences 18--22 are used as-is without augmentation, as they already contain authentic surgical degradations.

\begin{table}[t]
\centering
\scriptsize
\setlength{\tabcolsep}{4.5pt}
\caption{Robustness analysis under four train-test modes. 
$\Delta R$ = (Noisy$\to$Noisy) $-$ (Clean$\to$Noisy); 
$\Delta H$ = (Noisy$\to$Clean) $-$ (Clean$\to$Clean). Both computed on Dice.}
\label{tab:robustness}
\begin{tabular}{llccc cc}
\toprule
\textbf{Model} & \textbf{Mode} &
\textbf{Dice} & \textbf{IoU} & \textbf{Recall} &
\textbf{$\Delta R$} & \textbf{$\Delta H$} \\
\midrule

\multirow{4}{*}{\textbf{UNet\cite{ronneberger2015u}}}
 & Clean$\to$Clean & 0.8722 & 0.7757 & 0.8617 & \multirow{4}{*}{0.1548} & \multirow{4}{*}{-0.0234} \\
 & Clean$\to$Noisy & 0.6478 & 0.4848 & 0.6124 &  &  \\
 & Noisy$\to$Clean & 0.8488 & 0.7387 & 0.8575 &  &  \\
 & Noisy$\to$Noisy & 0.8026 & 0.6718 & 0.7954 &  &  \\
\midrule

\multirow{4}{*}{\textbf{SegFormer-B0\cite{xie2021segformer}}}
 & Clean$\to$Clean & 0.8971 & 0.8180 & 0.8617 & \multirow{4}{*}{0.1266} & \multirow{4}{*}{-0.0007} \\
 & Clean$\to$Noisy & 0.6962 & 0.5373 & 0.7345 &  &  \\
 & Noisy$\to$Clean & 0.8964 & 0.8147 & 0.8964 &  &  \\
 & Noisy$\to$Noisy & 0.8228 & 0.7021 & 0.8116 &  &  \\
\midrule

\multirow{4}{*}{\textbf{PraNet\cite{fan2020pranet}}}
 & Clean$\to$Clean & 0.9006 & 0.8214 & 0.8793 & \multirow{4}{*}{0.1279} & \multirow{4}{*}{-0.0164} \\
 & Clean$\to$Noisy & 0.7143 & 0.5625 & 0.7436 &  &  \\
 & Noisy$\to$Clean & 0.8842 & 0.7949 & 0.8687 &  &  \\
 & Noisy$\to$Noisy & 0.8422 & 0.7297 & 0.8494 &  &  \\
\midrule

\multirow{4}{*}{\textbf{CFFormer\cite{li2026cfformer}}}
 & Clean$\to$Clean & 0.9053 & 0.8293 & 0.9025 & \multirow{4}{*}{0.0846} & \multirow{4}{*}{-0.0152} \\
 & Clean$\to$Noisy & 0.7556 & 0.6089 & 0.7607 &  &  \\
 & Noisy$\to$Clean & 0.8901 & 0.8049 & 0.8947 &  &  \\
 & Noisy$\to$Noisy & 0.8402 & 0.7267 & 0.8491 &  &  \\
\midrule

\multirow{4}{*}{\textbf{\textsf{DepthPolyp} (Ours)}}
 & Clean$\to$Clean & 0.9107 & 0.8384 & 0.9038 & \multirow{4}{*}{0.0399} & \multirow{4}{*}{-0.0197} \\
 & Clean$\to$Noisy & 0.8126 & 0.6871 & 0.8129 &  &  \\
 & Noisy$\to$Clean & 0.8910 & 0.8054 & 0.8851 &  &  \\
 & Noisy$\to$Noisy & 0.8525 & 0.7447 & 0.8539 &  &  \\

\bottomrule
\end{tabular}
\end{table}

\begin{table}[b]
\centering
\scriptsize
\caption{Synthetic degradation specifications.}
\label{tab:degradation_spec}
\begin{tabular}{lcc}
\toprule
\textbf{Degradation Type} & \textbf{Parameters} & \textbf{Probability} \\
\midrule
Motion Blur & kernel: 3--29 px & 1.0 \\
Gaussian Blur & $\sigma \in \{3,5,7\}$ & 0.2 \\
Brightness & $\alpha \in [-0.1, 0.2]$ & 1.0 \\
Contrast & $\beta \in [-0.2, 0.2]$ & 1.0 \\
JPEG Compression & quality: 30--70 & 0.5 \\
Light Spots (reflection) & radius: 5--40 px, intensity: 0.85 & 0.8 \\
Fog & coef: 0.5--0.8 & 0.3 \\
Optical Distortion & distort/shift: 0.05 & 0.3 \\
\bottomrule
\end{tabular}
\end{table}

Table~\ref{tab:degradation_spec} summarizes the degradation pipeline. Motion blur and brightness/contrast adjustments model the most common surgical artifacts—camera shake from peristalsis and illumination instability. Light spots simulate specular reflections from wet mucosa. Each training sample is augmented to produce both clean and degraded versions, forcing the model to learn degradation-invariant structural features. This synthetic protocol accurately replicates real conditions, as validated by consistent improvements on authentic PolypGen degradations (Sec.~\ref{sec:polypgen_experiments}).

\subsection{Implementation Details}
\label{sec:implementation}

\noindent\textbf{Network Configuration.}
We adopt MiT-B0~\cite{xie2021segformer} as the encoder backbone. Input images are resized to $224 \times 224$ with standard augmentations including random horizontal flipping, color jittering, and the blur augmentations described in Sec.~\ref{sec:degradation}. Pseudo depth maps are generated using Depth-Anything v2-small~\cite{depth_anything_v2}.

\noindent\textbf{Training Protocol.}
All models are trained for 200 epochs using the AdamW optimizer with a learning rate $1 \times 10^{-4}$, weight decay $1 \times 10^{-4}$, with a warm-up strategy in first 10\% epochs and a cosine annealing strategy for learning rate adjustment. The batch size of 16 on an NVIDIA A100 GPU. 

\noindent\textbf{Inference Platforms.}
To assess deployment feasibility, we evaluate inference speed on: (1) NVIDIA RTX 3090 (FP32), (2) Apple iPhone 15 with CoreML (FP16), and (3) Raspberry Pi 4 (RPi 4), which is edge device with 4 cores A72 ARM-based CPU SoC.

\noindent\textbf{Evaluation Metrics.}
We report three standard segmentation metrics: Dice coefficient, Intersection over Union (IoU), and Recall. We emphasize Dice and IoU due to their clinical interpretability and widespread adoption in polyp segmentation benchmarks. For the complexity of models evaluation, we used the Multiply-Accumulate Operations (MACs).

\subsection{Robustness Four-quadrant Benchmark and Analysis}
\label{sec:robustness}

To systematically evaluate model robustness under realistic endoscopic degradations, we propose a four-quadrant benchmark that explicitly separates training and testing domain conditions. We define two domains: \textbf{Clean} (standard high-quality frames from Kvasir-SEG, CVC-ClinicDB, CVC-ColonDB) and \textbf{Noisy} (synthetically degraded samples plus authentic PolypGen sequences with motion blur, defocus, and specular reflections).

\noindent\textbf{Benchmark configurations.} 
We construct four train-test combinations to regard as the robustness validation: 
(1)~Clean$\to$Clean: baseline performance under matched conditions; 
(2)~Clean$\to$Noisy: generalization under distribution shift; 
(3)~Noisy$\to$Clean: clean-data performance cost; 
(4)~Noisy$\to$Noisy: upper-bound robustness with matched degradations.

Table~\ref{tab:robustness} presents robustness behavior across five representative architectures. \textbf{All models exhibit severe performance drops under Clean$\to$Noisy} (15.50\%--22.44\% Dice reduction), revealing that training on clean data fails to generalize to real surgical conditions. \textbf{Noisy training significantly recovers robustness}, as shown by $\Delta R$ = (Noisy$\to$Noisy) $-$ (Clean$\to$Noisy). \textsf{DepthPolyp} achieves the smallest robustness gap ($\Delta R = +0.0399$), outperforming UNet (+15.48\%) and PraNet (+12.79\%), demonstrating that depth-guided structural reasoning remains stable under severe degradation. \textbf{The clean-domain penalty is minimal} ($\Delta H < 2.2\%$ for all methods), with \textsf{DepthPolyp} at only $-0.0197$ Dice.

Since colonoscopy inherently involves motion blur, defocus, and reflections---conditions standard benchmarks ignore---\textbf{all subsequent experiments adopt noisy-trained models} to reflect realistic surgical deployment.

\begin{table}[!ht]
\centering
\caption{Cross-dataset generalization comparison under noise-aware training from heavyweight to lightweight model (up to down).
All models are trained on noisy Kvasir images and evaluated on clean (N→C) and noisy (N→N) test sets.
Results are reported as Dice$\uparrow$ / IoU$\uparrow$ / Recall$\uparrow$, where $\uparrow$ indicates higher is better.} 
\label{tab:final_comparison}
\resizebox{\textwidth}{!}{
\begin{tabular}{l c c l c c c}
\toprule
\textbf{Model} & \textbf{Params (M)} & \textbf{GMACs} &
\textbf{Eval.} &
\textbf{Kvasir} &
\textbf{Kvasir$\rightarrow$ClinicDB} &
\textbf{Kvasir$\rightarrow$ColonDB} \\
\midrule
\multicolumn{7}{c}{\textbf{Heavyweight CNN / Hybrid Models}} \\
\midrule
\multirow{2}{*}{NPDNet\cite{yu2024npd}} & \multirow{2}{*}{27.67} & \multirow{2}{*}{5.14}
 & N$\rightarrow$C & .845/.734/.855 & .765/.625/.752 & .697/.538/.717 \\
 & & 
 & N$\rightarrow$N & .804/.674/.814 & .681/.523/.673 & .563/.394/.644 \\
\midrule
\multirow{2}{*}{I2UNet-L\cite{dai2024i2u}} & \multirow{2}{*}{29.65} & \multirow{2}{*}{9.35}
 & N$\rightarrow$C & .837/.722/.840 & .706/.562/.642 & .644/.475/.642 \\
 & & 
 & N$\rightarrow$N & .799/.668/.819 & .591/.435/.568 & .606/.435/.610 \\
\midrule
\multirow{2}{*}{UNet\cite{ronneberger2015u}} & \multirow{2}{*}{31.04} & \multirow{2}{*}{41.93}
 & N$\rightarrow$C & .849/.739/.858 & .724/.579/.717 & .646/.482/.642 \\
 & & 
 & N$\rightarrow$N & .803/.672/.795 & .590/.432/.599 & .511/.347/.529 \\
\midrule
\multirow{2}{*}{PraNet\cite{fan2020pranet}} & \multirow{2}{*}{32.55} & \multirow{2}{*}{5.32}
 & N$\rightarrow$C & .884/.795/.869 & .832/.717/.810 & .765/.622/.703 \\
 & & 
 & N$\rightarrow$N & .842/.730/.849 & .678/.521/.688 & .650/.489/.646 \\
\midrule
\multirow{2}{*}{CTNet\cite{xiao2024ctnet}} & \multirow{2}{*}{44.29} & \multirow{2}{*}{6.27}
 & N$\rightarrow$C & .857/.751/.853 & .749/.603/.740 & .666/.505/.712 \\
 & & 
 & N$\rightarrow$N & .798/.666/.784 & .649/.486/.666 & .569/.403/.654 \\
\midrule
\multirow{2}{*}{SegFormer-B5\cite{xie2021segformer}} & \multirow{2}{*}{81.97} & \multirow{2}{*}{12.35}
 & N$\rightarrow$C & .889/.803/.893 & .865/.765/.830 & .823/.703/.825 \\
 & & 
 & N$\rightarrow$N & .850/.742/.862 & .757/.620/.737 & .725/.574/.753 \\
\midrule
\multirow{2}{*}{CFFormer\cite{li2026cfformer}} & \multirow{2}{*}{99.56} & \multirow{2}{*}{30.12}
 & N$\rightarrow$C & .890/.805/.895 & .851/.749/.843 & .766/.625/.772 \\
 & & 
 & N$\rightarrow$N & .840/.727/.849 & .730/.581/.753 & .662/.499/.648 \\
\midrule
\multicolumn{7}{c}{\textbf{Mid-size Models}} \\
\midrule
\multirow{2}{*}{I2UNet-S\cite{dai2024i2u}} & \multirow{2}{*}{7.03} & \multirow{2}{*}{2.73}
 & N$\rightarrow$C & .806/.677/.799 & .654/.496/.706 & .617/.455/.706 \\
 & & 
 & N$\rightarrow$N & .771/.629/.758 & .566/.411/.629 & .569/.401/.609 \\
\midrule
\multirow{2}{*}{CMUNeXt-L\cite{tang2024cmunext}} & \multirow{2}{*}{8.29} & \multirow{2}{*}{13.15}
 & N$\rightarrow$C & .776/.637/.750 & .642/.491/.622 & .596/.436/.610 \\
 & & 
 & N$\rightarrow$N & .761/.616/.766 & .568/.410/.565 & .535/.367/.670 \\
 \midrule
\multirow{2}{*}{H-Unets\cite{wu2024harmonizing}} & \multirow{2}{*}{16.22} & \multirow{2}{*}{12.78}
 & N$\rightarrow$C & .853/.746/.834 & .742/.599/.731 & .675/.511/.685 \\
 & & 
 & N$\rightarrow$N & .822/.699/.799 & .661/.499/.661 & .589/.426/.564 \\
\midrule
\multicolumn{7}{c}{\textbf{lightweight CNN / Hybrid Models}} \\
\midrule
\multirow{2}{*}{MobilePolypNet\cite{karmakar2022mobilepolyp}} & \multirow{2}{*}{0.22} & \multirow{2}{*}{0.96}
 & N$\rightarrow$C & .500/.335/.523 & .412/.269/.431 & .316/.189/.376 \\
 & & 
 & N$\rightarrow$N & .546/.377/.662 & .421/.271/.581 & .297/.176/.407 \\
\midrule
\multirow{2}{*}{ULite\cite{dinh2023ulite}} & \multirow{2}{*}{0.88} & \multirow{2}{*}{0.60}
 & N$\rightarrow$C & .759/.613/.745 & .647/.487/.637 & .507/.344/.505 \\
 & & 
 & N$\rightarrow$N & .727/.574/.744 & .584/.419/.593 & .480/.318/.533 \\
\midrule
\multirow{2}{*}{CMUNeXt-S\cite{tang2024cmunext}} & \multirow{2}{*}{0.42} & \multirow{2}{*}{0.83}
 & N$\rightarrow$C & .799/.667/.804 & .635/.478/.711 & .586/.426/.716 \\
 & & 
 & N$\rightarrow$N & .762/.618/.763 & .545/.395/.616 & .569/.406/.641 \\
\midrule
\multirow{2}{*}{CMUNeXt-B\cite{tang2024cmunext}} & \multirow{2}{*}{3.15} & \multirow{2}{*}{5.67}
 & N$\rightarrow$C & .756/.609/.738 & .636/.484/.575 & .481/.319/.511 \\
 & & 
 & N$\rightarrow$N & .738/.586/.774 & .571/.414/.563 & .522/.361/.549 \\
 \midrule
\multirow{2}{*}{UNeXt-L\cite{valanarasu2022unext}} & \multirow{2}{*}{1.47} & \multirow{2}{*}{0.44}
 & N$\rightarrow$C & .730/.576/.695 & .617/.461/.545 & .479/.320/.465 \\
 & & 
 & N$\rightarrow$N & .735/.584/.738 & .540/.382/.517 & .477/.318/.506 \\
\midrule
\multirow{2}{*}{MedT\cite{valanarasu2021medical}} & \multirow{2}{*}{1.56} & \multirow{2}{*}{1.80}
 & N$\rightarrow$C & .497/.333/.716 & .414/.267/.557 & .324/.198/.522 \\
 & & 
 & N$\rightarrow$N & .522/.355/.624 & .378/.237/.583 & .298/.179/.592 \\
\midrule
\multirow{2}{*}{SegFormer-B0\cite{xie2021segformer}} & \multirow{2}{*}{3.71} & \multirow{2}{*}{1.30}
 & N$\rightarrow$C & .896/.815/.896 & .830/.716/.832 & .763/.619/.719 \\
 & & 
 & N$\rightarrow$N & .823/.702/.812 & .698/.546/.688 & .621/.456/.590 \\
\midrule
\multirow{2}{*}{\textbf{\textsf{DepthPolyp} (Ours)}} & \multirow{2}{*}{\textbf{3.57}} & \multirow{2}{*}{\textbf{0.86}}
 & N$\rightarrow$C & .891/.805/.885 & .854/.748/.845 & .801/.669/.759 \\
 & & 
 & N$\rightarrow$N & \textbf{.853/.745/.854} & \textbf{.751/.608/.759} & \textbf{.734/.582/.697} \\
\bottomrule
\end{tabular}
}
\end{table}

\subsection{Comparison with State-of-the-Art Methods}

Table~\ref{tab:final_comparison} reports cross-dataset generalization results under noise-aware training, comparing \textsf{DepthPolyp} with 19 representative models ranging from ultra-lightweight to heavyweight architectures.

\textsf{DepthPolyp} achieves the best robustness under the N$\to$N setting across all test sets, obtaining Dice scores of 0.853, 0.751, 0.734 on Kvasir, ClinicDB, and ColonDB respectively. Compared with the strongest lightweight baseline SegFormer-B0 (3.71M parameters, 1.30 GMACs), our method consistently improves Dice (+3.6\%, +7.6\%, and +18.2\%, respectively) with 34\% less GMACs (0.86 vs.\ 1.30).

Among ultra-lightweight models ($<$1M parameters), ULite achieves the best
performance but remains substantially inferior to \textsf{DepthPolyp}
(0.727 vs.\ 0.853 Dice on Kvasir N$\to$N). In contrast, heavy models such as SegFormer-B5 achieve competitive clean performance but exhibit noticeable degradation under noisy conditions despite one to two orders of magnitude higher computational cost.

Notably, \textsf{DepthPolyp} shows a small gap between N$\to$C and N$\to$N
performance (average Dice drop of 4.9\%), indicating stable behavior across
varying image quality and highlighting the benefit of depth-guided structural feature fusion for real-world deployment.

\subsection{Real-World Deployment: PolypGen Evaluation and Inference Efficiency}
\label{sec:polypgen_experiments}

\begin{figure}[b]
\centering
\includegraphics[width=0.95\textwidth]{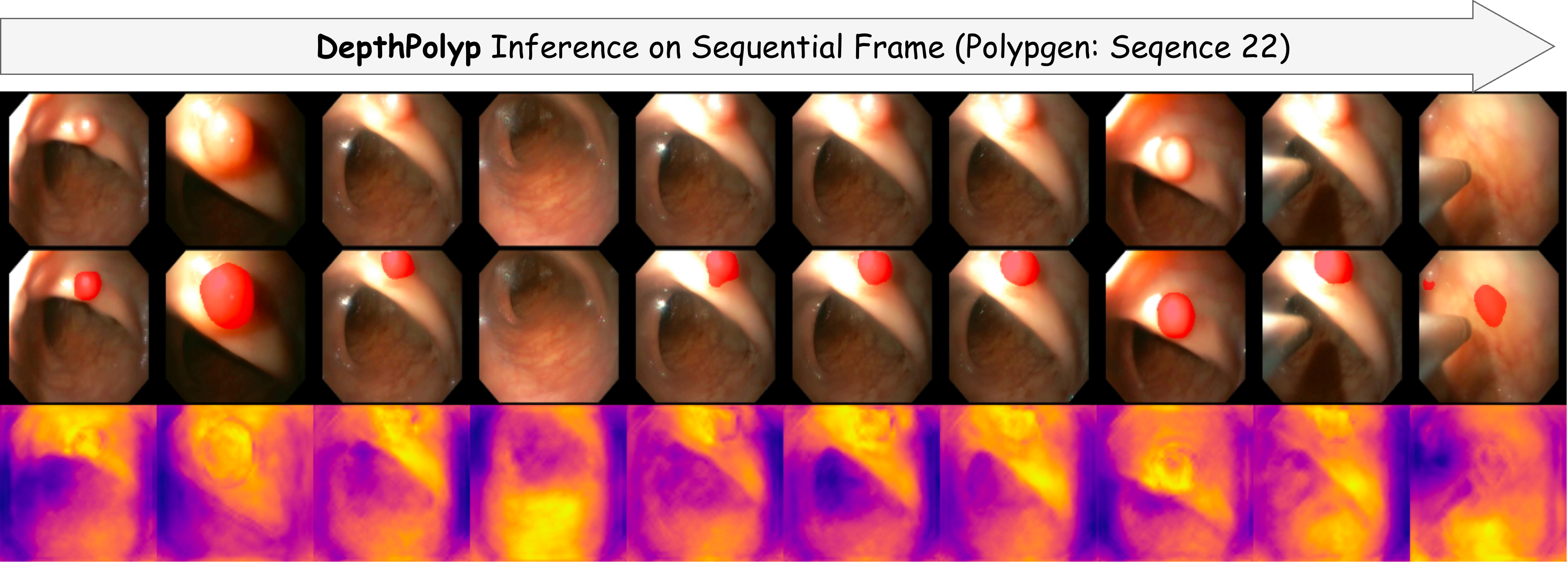}
\caption{Qualitative results of \textsf{DepthPolyp} on sequential PolypGen frames (Sequence~22), showing input images, predicted polyp masks, and depth-aware representations.}
\label{fig:polypgen_quali}
\end{figure}


While cross-dataset benchmarks evaluate generalization under controlled settings,
real-world deployment further requires robustness to severe surgical degradations
and real-time inference efficiency.
We evaluate representative models on PolypGen sequences 18--22, which contain
authentic artifacts such as motion blur, defocus, and specular reflections.
Table~\ref{tab:polypgen_inference} reports average cross-dataset Dice (mean N$\to$N
performance on Kvasir, ClinicDB, and ColonDB), PolypGen Dice, and inference speed
measured at batch size 1 to simulate real-time video processing. 

\textsf{DepthPolyp} achieves the best performance on PolypGen, obtaining
0.779 average Dice and 0.679 PolypGen Dice, outperforming all lightweight models by a large margin. Compared with SegFormer-B0, the strongest lightweight baseline, our method improves average Dice by +9.1\% and PolypGen Dice by +7.1\%. Notably, \textsf{DepthPolyp} also surpasses heavyweight models such as CFFormer (99.56M parameters, 30.12 GMACs) on PolypGen while using over 96\% fewer parameters and operations. The inference results of \textsf{DepthPolyp} on PolypGen Sequence 22 are shown in Fig.~\ref{fig:polypgen_quali}. \textsf{DepthPolyp} provides robust segmentation and relative depth information, aiding more accurate distance measurement during surgery.

In terms of efficiency, SegFormer-B5 achieves comparable average Dice (0.777) but requires 23$\times$ more parameters and 14$\times$ more computation. Ultra-lightweight models (e.g., CMUNeXt-S and ULite) exhibit severe performance degradation on PolypGen, indicating that aggressive parameter reduction is insufficient for handling complex surgical artifacts.

Finally, real-time inference results on NVIDIA RTX 3090, Apple iPhone~15, and Raspberry RPi~4 demonstrate that \textsf{DepthPolyp} supports practical deployment across workstation, mobile, and embedded platforms, enabled by its low computational cost of 0.86~GMACs.

\begin{table}[t]
\centering
\scriptsize
\setlength{\tabcolsep}{3pt}
\caption{Real-world robustness and inference efficiency. Average Dice is computed across Kvasir, ClinicDB, and ColonDB under N$\to$N evaluation. Inference speed (FPS) is measured with batch size 1.}
\label{tab:polypgen_inference}
\begin{tabular}{lccccccccc}
\toprule
\textbf{Model} & \textbf{Params} & \textbf{GMACs} & \textbf{Avg.} & \multicolumn{3}{c}{\textbf{PolypGen}} & \multicolumn{3}{c}{\textbf{FPS (Batch=1)}} \\
\cmidrule(lr){5-7} \cmidrule(lr){8-10}
 & \textbf{(M)} &  & \textbf{Dice} & \textbf{Dice} & \textbf{IoU} & \textbf{Rec.} & \textbf{GPU} & \textbf{iPhone} & \textbf{RPi 4} \\
\midrule
\multicolumn{10}{c}{\textit{Heavyweight Models ($>$30M)}} \\
\midrule
NPDNet\cite{yu2024npd} & 27.67 & 5.14 & .683 & .542 & .462 & .747 & 28.41 & 63.49 & 1.17 \\
UNet\cite{ronneberger2015u} & 31.04 & 41.93 & .635 & .579 & .510 & .706 & 139.47 & 36.51 & 0.33 \\
PraNet\cite{fan2020pranet} & 32.55 & 5.32 & .723 & .676 & .605 & .708 & 58.08 & 71.53 & 1.83 \\
SegFormer-B5\cite{xie2021segformer} & 81.97 & 12.35 & .777 & .671 & .609 & .770 & 17.22 & 41.05 & 0.56 \\
CFFormer\cite{li2026cfformer} & 99.56 & 30.12 & .744 & .643 & .584 & .767 & 34.97 & 21.72 & 0.42 \\
\midrule
\multicolumn{10}{c}{\textit{Mid-size Models (5--20M)}} \\
\midrule
I2UNet-S\cite{dai2024i2u} & 7.03 & 2.73 & .635 & .433 & .350 & .674 & 69.10 & 90.25 & 2.22 \\
CMUNeXt-L\cite{tang2024cmunext} & 8.29 & 13.15 & .621 & .443 & .370 & .651 & 80.84 & 69.66 & 0.24 \\
H-Unets\cite{wu2024harmonizing} & 16.22 & 12.78 & .691 & .550 & .472 & .655 & 29.07 & 75.42 & 0.64 \\
\midrule
\multicolumn{10}{c}{\textit{Lightweight Models ($<$5M)}} \\
\midrule
ULite\cite{dinh2023ulite} & 0.88 & 0.60 & .597 & .387 & .305 & .582 & 142.68 & 285.13 & 5.32 \\
MedT\cite{valanarasu2021medical} & 1.56 & 1.80 & .400 & .199 & .130 & .491 & 3.27 & 3.17 & 0.14 \\
CMUNeXt-B\cite{tang2024cmunext} & 3.15 & 5.67 & .610 & .504 & .423 & .664 & 110.25 & 141.64 & 0.60 \\
SegFormer-B0\cite{xie2021segformer} & 3.71 & 1.30 & .714 & .634 & .579 & .805 & 84.14 & 186.72 & 4.12 \\
\textbf{\textsf{DepthPolyp}(Ours)} & \textbf{3.57} & \textbf{0.86} & \textbf{.779} & \textbf{.679} & \textbf{.620} & .788 & 79.12 & 181.54 & 4.05 \\
\bottomrule
\end{tabular}
\end{table}

\subsection{Ablation Study}


\begin{table}[t]
\centering
\scriptsize
\setlength{\tabcolsep}{4pt}
\caption{Ablation study on key components. All variants are trained and evaluated under N$\to$N protocol. Avg. Dice and Recall are computed across Kvasir, ClinicDB, and ColonDB.}
\label{tab:ablation}
\begin{tabular}{lccccc}
\toprule
\textbf{Model Variant} & \textbf{Params} & \textbf{GMACs} & \textbf{Avg.} & \textbf{Avg.} & \textbf{FPS} \\
 & \textbf{(M)} &  & \textbf{Dice} & \textbf{Recall} & \textbf{(iPhone)} \\
\midrule
\textbf{\textsf{DepthPolyp} (Full)} & \textbf{3.57} & \textbf{0.86} & \textbf{.784} & \textbf{.807} & \textbf{181.54} \\
\midrule
w/o Depth Guidance & 3.57 & 0.86 & .759 & .789 & 181.54 \\
w/o Uncertainty Loss & 3.57 & 0.86 & .605 & .674 & 181.54 \\
w/o GFM & 3.73 & 1.36 & .776 & .798 & 131.39 \\
w/o ISF & 3.57 & 0.84 & .760 & .780 & 169.91 \\
w/o DGG & 3.57 & 0.86 & .736 & .752 & 147.87 \\
\bottomrule
\end{tabular}
\end{table}

We perform systematic ablations to quantify the contribution of each component (Table~\ref{tab:ablation}).
Removing the uncertainty-weighted loss leads to the most severe performance degradation, with Dice dropping from 0.784 to 0.605. The observed performance degradation without uncertainty weighting implies its relevance for multi-task optimization stability.
Architectural ablations show complementary effects: removing DGG reduces Dice to 0.736 and lowers iPhone throughput to 147.9 FPS; removing the ISF yields Dice 0.760 with a modest speed drop to 169.9 FPS; removing GFM has minor Dice impact (0.776) but substantially reduces speed (131.4 FPS), indicating its primary role in efficiency. Omitting depth guidance produces a moderate Dice decrease to 0.759, suggesting pseudo-depth acts as a useful structural regularizer. Overall, the full \textsf{DepthPolyp} model attains the best trade-off between accuracy and runtime, with each module contributing uniquely to robustness, efficiency, or both.

\subsection{Qualitative Results}
\begin{figure}[!t]
\includegraphics[width=\textwidth]{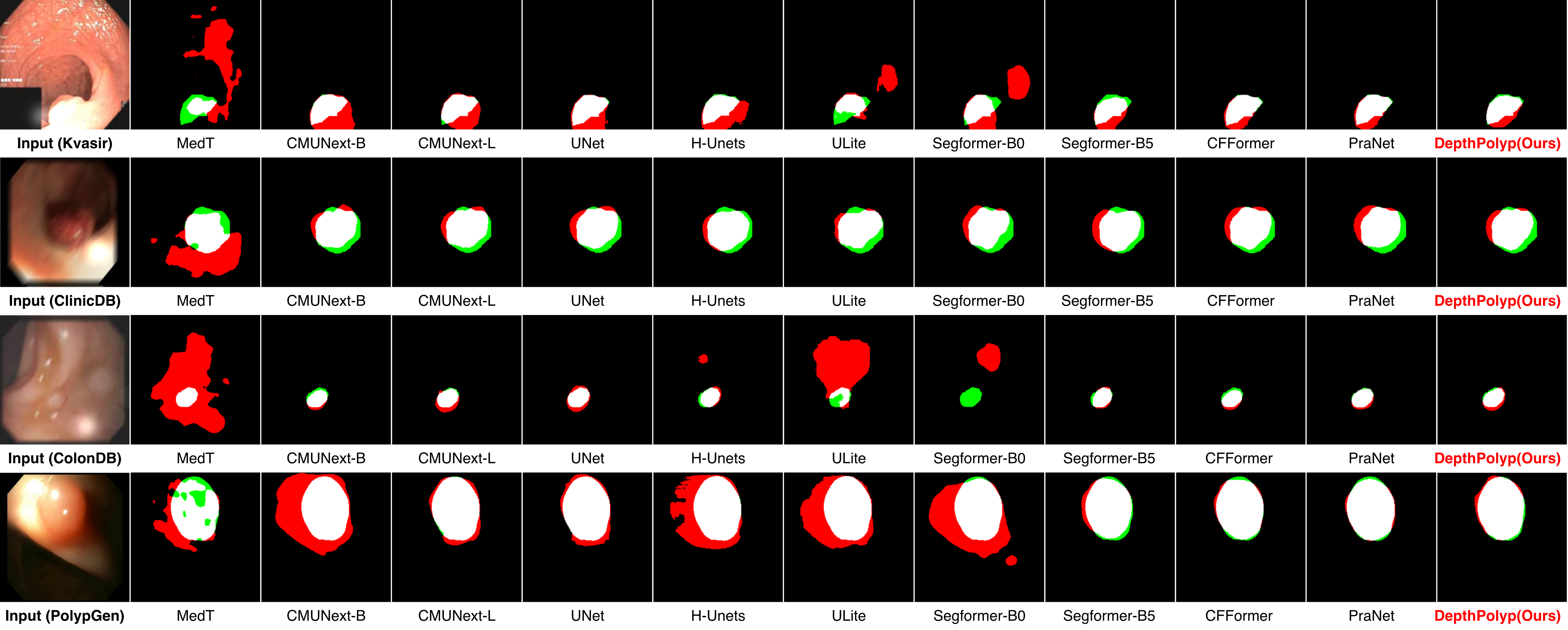}
\caption{Qualitative comparison on challenging colonoscopy images affected by motion blur, illumination variation, low contrast, and specular highlights. 
Each row corresponds to one test case. From left to right, the columns show the input image, reference annotation, predictions from representative baseline methods, and \textsf{DepthPolyp (Ours)}. 
White denotes true positives, red false positives, and green false negatives.}

\label{fig:qualitative}
\end{figure}
Fig.~\ref{fig:qualitative} presents representative qualitative results on challenging colonoscopy images with severe degradations, including motion blur, uneven illumination, specular highlights, and low contrast. Compared with representative baselines, \textsf{DepthPolyp} produces more compact and anatomically coherent segmentations, with clearer boundaries and substantially fewer false positives.

Many baseline methods exhibit fragmented predictions, boundary leakage, or spurious responses when appearance cues are unreliable, especially for small polyps or reflective regions. In contrast, \textsf{DepthPolyp} maintains stable localization and suppresses isolated false activations across diverse degradation patterns.

These visual comparisons are consistent with the quantitative improvements reported in Table~\ref{tab:final_comparison} and~\ref{tab:polypgen_inference}, highlighting the robustness of pseudo-depth-guided and uncertainty-aware learning under real-world surgical conditions.

\vspace{-2mm}

\section{Conclusion}
This paper identifies robustness as a critical limitation of existing polyp segmentation models in real endoscopic scenarios, where performance often degrades under blur, noise, and illumination variation despite strong clean-test results. To address this issue, we propose \textsf{DepthPolyp}, a lightweight framework that improves robustness without increasing model size or computational cost. Extensive experiments show that stable performance is achieved through the joint use of pseudo-depth guidance and model design, rather than increased model capacity.

Beyond performance gains, our analysis suggests that pseudo-depth supervision primarily acts as a training regularizer, guiding models toward degradation-tolerant representations. Moreover, the robustness-oriented evaluation protocol adopted in this work exposes failure modes overlooked by conventional clean benchmarks, highlighting the need for more deployment-focused design and evaluation in future medical image segmentation research.






%
%
%

{\small
\bibliographystyle{splncs04}
\bibliography{software}
}

%






\end{document}